\documentclass[journal]{IEEEtran}

 \usepackage[caption=false,font=footnotesize,labelfont=sf,textfont=sf]{subfig}

 \usepackage[caption=false,font=footnotesize]{subfig}

\usepackage{graphicx}
\usepackage{amsmath} 
\usepackage{amssymb}
\usepackage{algorithmic} 
\usepackage{array}
\usepackage{url}  
\usepackage{color}
\usepackage{multirow}
\usepackage[colorlinks, linkcolor=blue, anchorcolor=blue, citecolor=blue]{hyperref}
\usepackage[numbers]{natbib}
\usepackage{gensymb}
\usepackage{url}
\usepackage{colortbl}
\usepackage{booktabs}
\usepackage{tabularx}
\usepackage{bbding}
\usepackage{threeparttable}
\usepackage{changes}

\newcommand{\ie}{i.e.}

\begin{document}
%
\title{Vision-Language Navigation:\\ A Survey and Taxonomy}
%
%
%

\author{ Wansen~Wu,
	Tao~Chang,
	Xinmeng~Li,
	Quanjun~Yin,
	Jiancheng~Zhu
	
\thanks{Manuscript received xx xx, 2020; revised xx xx, 2020.}
\thanks{W. Wu, X. Li, Q. Yin, and J. Zhu (the corresponding author) are with the College of Systems Engineering, National University of Defense Technology, Changsha, Hunan, 410073 China (e-mails: wuwansen14@nudt.edu.cn; xml.nudt@gmail.com; yin\_quanjun@163.com; zhujiancheng2012@126.com). 
}
\thanks{Tao~Chang was with National Key Laboratory of Parallel and Distributed Processing, College of Computer Science and Technology, National University of Defence Technology, Changsha 410073, China (changtao15@nudt.edu.cn).}
}

%
%

\markboth{IEEE Transactions on Neural Networks and Learning Systems,~Vol.~x, No.~x, June~2021}%
{Wu \MakeLowercase{\textit{et al.}}: Vision-and-Language Navigation: A Survey and Taxonomy}
%



\maketitle

\begin{abstract}
	Vision-Language Navigation (VLN) tasks require an agent to follow human language instructions to navigate in previously unseen environments. This challenging field involving problems in natural language processing, computer vision, robotics, etc., has spawn many excellent works focusing on various VLN tasks. This paper provides a comprehensive survey and an insightful taxonomy of these tasks based on the different characteristics of language instructions in these tasks. Depending on whether the navigation instructions are given for once or multiple times, this paper divides the tasks into two categories, i.e., \textit{single-turn} and \textit{multi-turn} tasks.  For single-turn tasks, we further subdivide them into \textit{goal-oriented} and \textit{route-oriented} based on whether the instructions designate a single goal location or specify a sequence of multiple locations. For multi-turn tasks, we subdivide them into \textit{passive} and \textit{interactive} tasks based on whether the agent is allowed to question the instruction or not. These tasks require different capabilities of the agent and entail various model designs. We identify progress made on the tasks and look into the limitations of existing VLN models and task settings. Finally, we discuss several open issues of VLN and point out some opportunities in the future, i.e., incorporating knowledge with VLN models and implementing them in the real physical world.

\end{abstract}


%
\IEEEpeerreviewmaketitle

\section{Introduction} \label{sec:intro}

\begin{figure}[tb]
	\centering
	\includegraphics[width=1.1\linewidth]{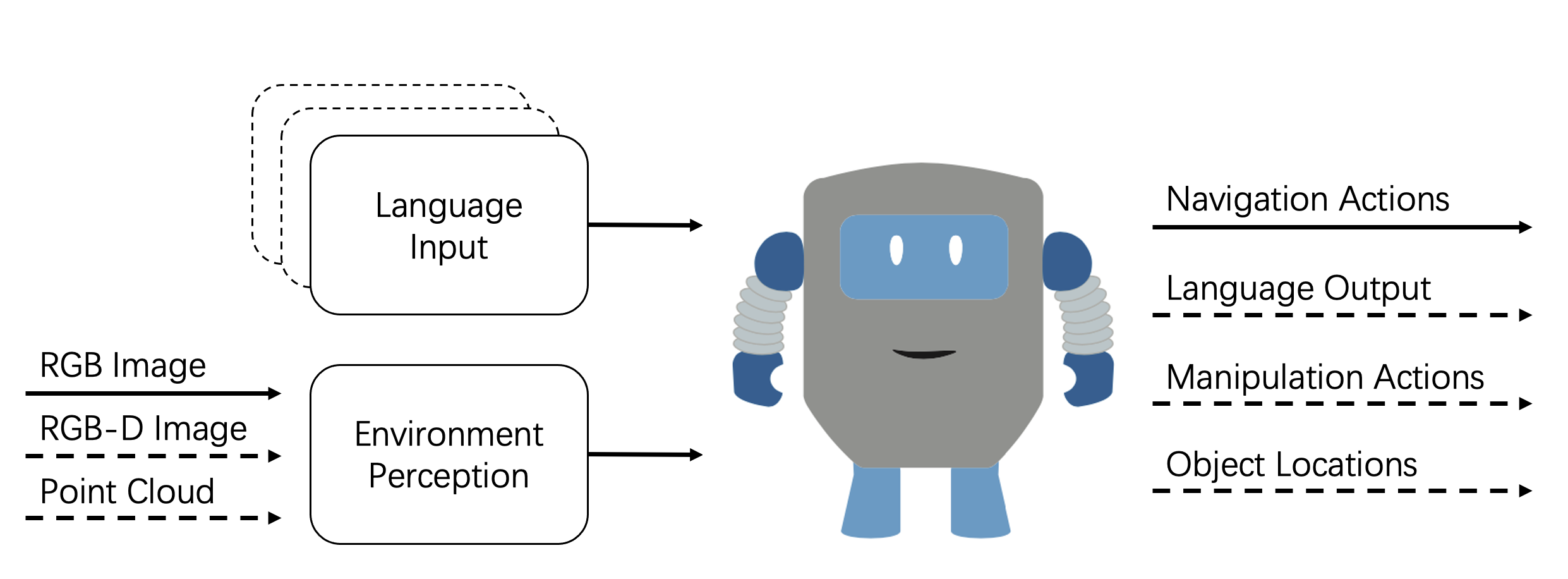}
	\caption{System diagram showing input and output of an agent linking vision-and-language to action. The solid line in/output modules are essential for a Visual-and-Language Navigation agent.}
	\label{fig:diagram}
\end{figure}

Artificial Intelligent systems should be able to sense the environment and make actions in order to perform specific tasks~\cite{russell2002artificial}. Computer Vision (CV) and Natural Language Processing (NLP) as the main technologies for agents to perceive the environment, have also received extensive attention from researchers. Significant progress have been made in CV and NLP fields largely due to the successful application of deep learning such as classification~\citep{DBLP:conf/cvpr/HeZRS16}, object detection~\citep{DBLP:conf/cvpr/RedmonF17}, segmentation~\citep{DBLP:journals/pami/HeGDG20}, large-scale pre-trained language model~\citep{DBLP:conf/naacl/DevlinCLT19,DBLP:conf/nips/ConneauL19},
etc. Deep learning has also achieved certain results in the field of robot control and decision-making, such as \cite{DBLP:journals/scirobotics/LeeDBTKH19}. These tremendous advances have fueled the confidence of researchers for solving more complex tasks that involve language embedding, vision encoding and decision-making on actions. Vision-Language Navigation (VLN) is this type of task, which requires agent to follow human language instructions to navigate in previously unseen environments. VLN task is more in line with human expectations of artificial intelligence, and has inspired a series of subsequent work, such as manipulation using vision and language~\citep{DBLP:conf/cvpr/ShridharTGBHMZF20} and outdoor navigation~\citep{DBLP:journals/corr/abs-2002-11310}. As shown in Figure \ref{fig:diagram}, a VLN agent typically takes visual observations and language instructions as inputs and output is either navigation actions, natural language, manipulation actions or identifying object locations. 

\begin{center}
	\begin{figure*}[tb]
		\centering
		\includegraphics[width=0.75\linewidth]{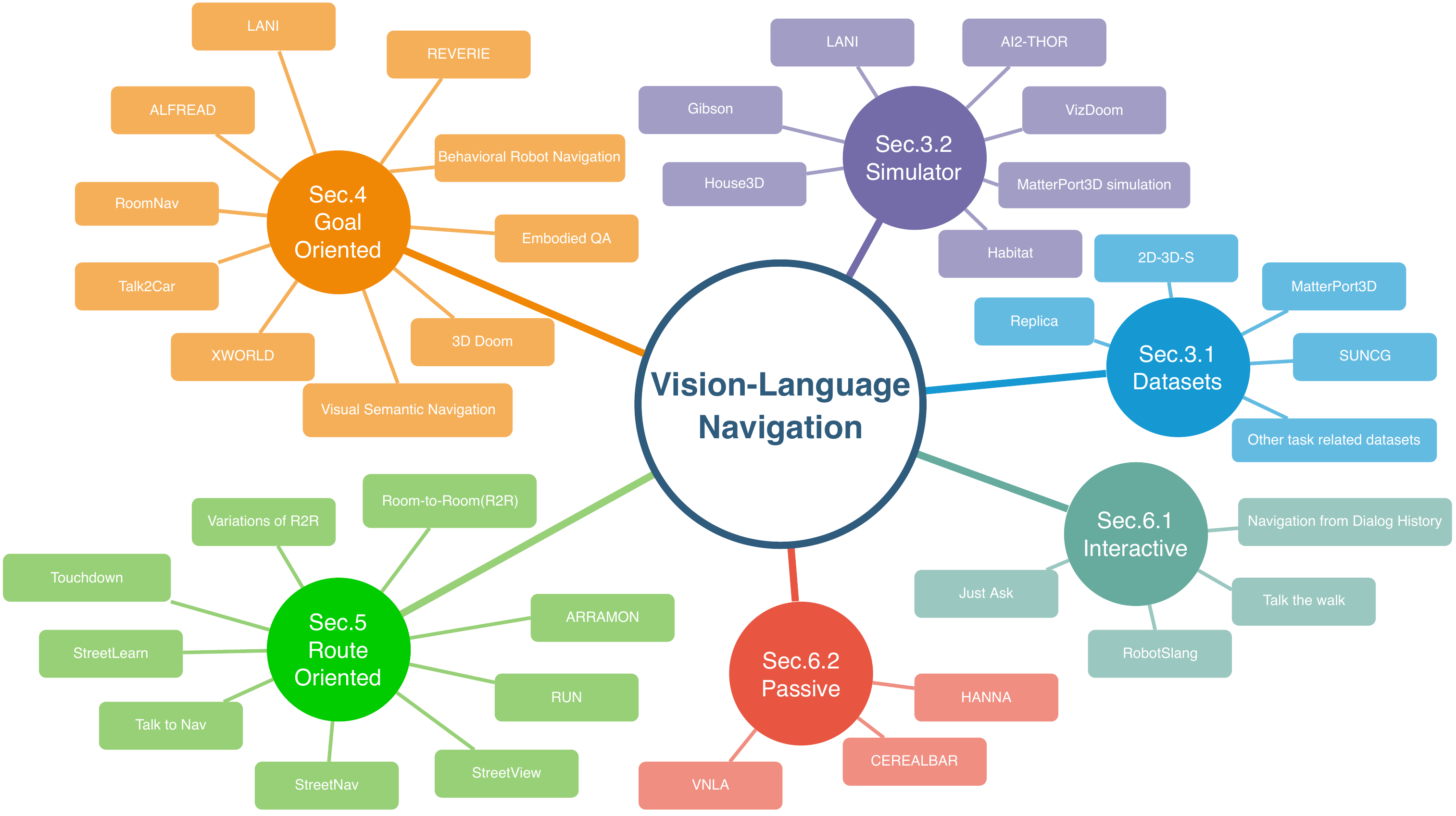}
		\caption{The knowledge graph is summarized in this review.}
		\label{fig:arch}
	\end{figure*}
\end{center}

In this survey, we will comprehensively review existing vision-and-language literatures for different tasks, not only the task proposed by ~\citet{DBLP:conf/cvpr/AndersonWTB0S0G18} in 2018, but also other tasks that utilize both natural language and visual information for navigation. Futhermore, we try to find out the inherent differences among the VLN tasks. We found that the visual input of the same type of task is basically similar, i.e., environments for indoor tasks are mostly home, while for outdoor tasks are street scenes. There may also be different task settings in the same environment, so these tasks cannot be effectively classified by visual input. For example, REVERIE~\citep{DBLP:conf/cvpr/QiW0WWSH20} task and Room-to-Room~\cite{DBLP:conf/cvpr/AndersonWTB0S0G18} task are based on the same environment simulator, but they have different task settings. In comparison, the language part leads these tasks completely different. Some tasks give specific instructions when the agent navigation starts, and the form of the instructions varies. Some tasks require agents to interact with humans to obtain information progressively, so as to complete the navigation task. Based on this observation, we introduce a new taxonomy to categorize existing different VLN tasks.  Depending on navigation instruction is given for once or multiple times, the VLN tasks can be divided into two types: \textit{single-turn} and \textit{multi-turn} tasks. 

For \textbf{single-turn} tasks, the agent receives a language instruction at start position and then navigates to the destination following the instruction. Depending on whether the specific route is described in the instructions, single-turn task can be subdivided into: \textit{Goal-oriented} tasks and \textit{Route-oriented} tasks. 
\begin{itemize}
	\item \textbf{Goal-oriented}. \added{The instructions contain one or more goals, but does not give detailed routes to reach these goals. The goals are often visible at the initial position, but there are also invisible situations, which require the agent to find in the environment by moving or interacting with the environment(e.g., ``bring me a spoon", spoon is in the drawer, only can be found by openning the drawer). After finding the target, the agent needs to plan the  trajectory by itself to reach the target position.}
	\item \textbf{Route-oriented}. In this type of tasks, the language instruction describes the objects seen along the way and the agent's route in detail. The agent may get lost if it does not strictly follow the route indicated in the instruction which is usually well-formed. Instructions can be decomposed into several meaningful pieces by some rules and each of them indicates an movement action. In this case, the agent can plan an action sequence, then carries out them. 
\end{itemize}

For \textbf{multi-turn} tasks, instructions will be given by a guide to VLN agent in several turns, until the agent has reached the specified goal. According to whether the agent can question the instruction or not, we subdivided multi-turn tasks into: \textit{Passive} and \textit{Interactive} tasks. \added{As different instructions can be given sequentially, an instruction in a single turn is mainly goal-oriented for easy execution, where the goal is usually visible when the agent is required to perform the instruction.} 
\begin{itemize}
	\item \textbf{Passive}. Instructions are given to the agent in stages, where the information contained in the instructions is often sufficient and unambiguous, so the agent only needs to understand the meanings and move to the target position. Order instructions for this type of task is different for this problem. 
	\item \textbf{Interactive}. In this task setting, the human guide and the agent are usually cooperative, and the agent can ask the person if it encounters a situation of information insufficiency or ambiguity. In this interactive manner, the agent can continuously acquire information to complete the navigation task.
\end{itemize}

\textbf{Scope of the survey.} Existing reviews related to VLN are mainly in Embodied AI~\cite{pfeifer2004embodied,pfeifer2006body,duan2022survey} areas and multi-modal research in vision and language~\cite{baltruvsaitis2018multimodal,uppal2022multimodal}. Other related paper like~\cite{kruse2013human} reviews previous works of traditional navigation model based on geometrical and topological approaches, and~\citet{guilherme2002vision} reviews the robotic navigation literature. Review papers mentioned above are related to VLN approaches, but none of the reviews specifically studied the VLN task. Since VLN has now become a popular field, models and datasets are also being updated very quickly. Therefore, a comprehensive review of the progress and taxonomy of related tasks that enable researchers to better grasp the key point of a specific task and identify directions for future research is necessary. Due to limitations on space and our knowledge, we apologize to those authors whose works are not included in this paper.

\begin{table*}[htb]
	\centering 
	\caption{Taxonomies and statics of VLN tasks. Beyond navigation, a task may interleave with other actions, the M, Q, L in column \textbf{Compound} mean that an agent is required to manipulate an object, answering an question, locate a target object, respectively. The Matterport3D in column \textbf{Simulator} means Matterport3D Simulator. The - in column \textbf{Simulator} indicates the name of simulator is the same as the task, or not named.}
	\begin{threeparttable}
		\begin{tabular}{c|c|c|cccc}
			\toprule 
			\textbf{Task Type} & \multicolumn{2}{c|}{\textbf{Instruction}} & \textbf{Name} & \textbf{Simulator} & \textbf{Outdoor} & \textbf{Compound} \\
			\midrule
			\multirow{18}{*}{Single} &  \multicolumn{2}{c|}{\multirow{11}{*}{Goal-Oriented} } & LANI~\cite{DBLP:conf/emnlp/MisraBBNSA18} & - &  & - \\ 
			& \multicolumn{2}{c|}{~} & ALFRED~\citep{DBLP:conf/cvpr/ShridharTGBHMZF20} & AI2-THOR~\citep{DBLP:journals/corr/abs-1712-05474} &   & M \\ 
			& \multicolumn{2}{c|}{~}   & Talk2Car~\citep{DBLP:conf/emnlp/DeruyttereVGGM19} & - & {\color{red} \CheckmarkBold} & M \\ 
			& \multicolumn{2}{c|}{~}  & XWORLD~\citep{DBLP:conf/corl/YuLZX18} & - &  & - \\ 
			&  \multicolumn{2}{c|}{~}   & 3D Doom\citep{DBLP:conf/aaai/ChaplotSPRS18} & VizDoom~\citep{DBLP:conf/cig/KempkaWRTJ16} &  & - \\ 
			&  \multicolumn{2}{c|}{~}   & Visual Semantic Navigation & AI2-THOR~\citep{DBLP:journals/corr/abs-1712-05474} &   & \\ \cline{4-7}
			
			&    \multicolumn{2}{c|}{~} &  EQA~\citep{DBLP:conf/cvpr/DasDGLPB18} & House3D~\citep{DBLP:conf/iclr/WuWGT18} &   & Q \\ 
			&    \multicolumn{2}{c|}{~}   & RoomNav~\citep{DBLP:conf/iclr/WuWGT18} & House3D~\citep{DBLP:conf/iclr/WuWGT18} &    & -  \\ 
			&   \multicolumn{2}{c|}{~} & REVERIE~\citep{DBLP:conf/cvpr/QiW0WWSH20} & Matterport3D~\citep{DBLP:conf/cvpr/AndersonWTB0S0G18} &   & L \\ 
			&    \multicolumn{2}{c|}{~}   & Behavioral Robot Navigation\citep{DBLP:conf/emnlp/ZangPVCNSS18} & - &   & - \\ 
			&     \multicolumn{2}{c|}{~}  & Navigation Task Based on SUNCG\citep{DBLP:conf/iclr/FuKLG19} & - &   &  - \\ \cline{2-7}
			
			& \multicolumn{2}{c|}{\multirow{11}*{Route-Oriented}} & Room-to-Room~\cite{DBLP:conf/cvpr/AndersonWTB0S0G18} & Matterport3D &   & - \\ 
			& \multicolumn{2}{c|}{~} & Room-for-Room~\citep{DBLP:conf/acl/JainMKVIB19} & Matterport3D &   & - \\ 
			& \multicolumn{2}{c|}{~} & Room-Cross-Room~\cite{DBLP:conf/emnlp/KuAPIB20} & Matterport3D&   & -\\ 
			& \multicolumn{2}{c|}{~} & R6R, R8R~\cite{zhu2020babywalk} & Matterport3D &   & -\\ 
			& \multicolumn{2}{c|}{~} & Room-to-Room-CE~\cite{DBLP:conf/eccv/KrantzWMBL20} & Matterport3D&  & -\\ 
			& \multicolumn{2}{c|}{~} & Cross lingual Room-to-Room~\cite{DBLP:journals/corr/abs-1910-11301} & Matterport3D&   & -\\ 
			& \multicolumn{2}{c|}{~} & TouchDown~\cite{DBLP:journals/corr/abs-1811-12354} & Google Street View \tnote{1} & {\color{red} \CheckmarkBold} & L\\ 
			& \multicolumn{2}{c|}{~} & RUN~\cite{DBLP:conf/emnlp/Paz-ArgamanT19} & - &  & -\\
			& \multicolumn{2}{c|}{~} & Street Nav~\cite{DBLP:conf/aaai/HermannMMBAH20} & Google Street View & {\color{red} \CheckmarkBold} & -\\ 
			& \multicolumn{2}{c|}{~} & StreetLearn~\cite{DBLP:journals/corr/abs-1903-01292} & Google Street View & {\color{red} \CheckmarkBold} & -\\ 
			& \multicolumn{2}{c|}{~} & ARRAMON~\cite{DBLP:conf/emnlp/KimZBTB20} & - &  & M \\ \hline
			
			\multirow{7}{*}{Multi} & \multicolumn{2}{c|}{\multirow{3}*{Passive}} & CEREALBAR~\citep{DBLP:conf/emnlp/SuhrYSYKMZA19}& - &  &  - \\ 
			& \multicolumn{2}{c|}{~} & VLNA~\citep{DBLP:conf/cvpr/NguyenDBD19} & Matterport3D &   & - \\ 
			& \multicolumn{2}{c|}{~} & HANNA~\citep{DBLP:conf/emnlp/NguyenD19} & Matterport3D &   & - \\ \cline{2-7}
			
			& \multicolumn{2}{c|}{\multirow{4}*{Interactive}} & CVDN~\citep{DBLP:conf/corl/ThomasonMCZ19} & Matterport3D&  & -  \\ 
			& \multicolumn{2}{c|}{~} & Just Ask\citep{chi2020just} &  Matterport3D &   & - \\
			& \multicolumn{2}{c|}{~} & Talk The Walk~\citep{DBLP:journals/corr/abs-1807-03367} & - & { \color{red} \CheckmarkBold} & - \\ 
			& \multicolumn{2}{c|}{~} & RobotSlang~\citep{DBLP:journals/corr/abs-2010-12639} & Physical &   & - \\ 
			\bottomrule
		\end{tabular}
	\end{threeparttable}
	\begin{tablenotes}
		\item[1] \url{https://developers.google.com/maps/documentation/streetview/intro}
	\end{tablenotes}
	\label{tab:tax}
\end{table*}

\textbf{Contributions.} First, to the best of our knowledge, this paper is the first work to give a comprehensive review of the existing vision-and-language tasks . Compared to recent works, this survey covers more recent papers, more tasks and datasets. Second, the taxonomy of VLN tasks is well-designed, it can help researchers better understand the task. Third, limitations of current approaches are fully described and discussed. We also proposed some possible solutions, which may provide inspiring ideas for researchers in this field.

The remainder of this paper is organized as follows: In Section II, we briefly provide some technical background in CV, NLP and robot navigation field. Then, in Section III, different types of VLN dataset and simulator are described in detail. As shown in Tab.~\ref{tab:tax}, we summarize typical datasets and related simulators, the tasks will be illustrated in Section IV. Beyond navigation, some tasks may interleave with other actions, such as manipulate an object, answer an question, locate a target object. Note that our taxonomy only considers the navigation part if a task have multiple task settings. Finally, the conclusions are presented in Section V. A knowledge graph is summarized in Figure~\ref{fig:arch}. This can help readers to quickly find content of interest.

\section{Preliminaries} \label{sec:bg}
In this section, we briefly provide technical background needed of VLN tasks. We review the effect of deep learning to computer vision, natural language processing, robot navigation and Visual-Linguistic Learning, which are preliminaries of VLN tasks.

\subsection{Computer Vision}
The advent of deep learning~\citep{DBLP:journals/nature/LeCunBH15} has tremendously changed the field of computer vision. The best way to represent images is by leveraging automatic feature extraction methods. Convolutional Neural Networks (CNNs)~\citep{lecun1998gradient} have become the de facto standard for generating representations of images using end-to-end trainable models. Residual Networks~\citet{DBLP:conf/cvpr/HeZRS16} make the layer number of the network increase from a dozen to dozens, or even more than 100. 

With these novel architecture, all task of computer vision, such as Image Classification, Object Localization, Object Detection, Object Segmentation, Object Identification, Instance segmentation and Panoptic segmentation, have made tremendous progress. 

\subsection{Statistical Natural Language Processing}
Language is usually represented either with bag-of-words or with sentence representations. Modern approaches use word vectors to capture or learn structure, such as Long Short-Term Memory units (LSTMs)~\citep{DBLP:journals/neco/HochreiterS97} combined with Word2Vec~\citep{DBLP:journals/corr/abs-1301-3781} or Paragraph Vector~\citep{DBLP:conf/icml/LeM14}. These approaches learn a vector representation associated with either words or longer documents and then compute over an entire sentence to perform tasks such as shallow parsing, syntax parsing, semantic role labeling, named entity recognition, entity linking, co-reference resolution, etc.

\subsection{Robot Navigation}
Navigation is a important ability for artificial agents to adapt to an environment and is the precondition for other advanced behaviors. Robot navigation task expects robot find optimal path to reach the destination, typical navigation tasks including standard vision-based navigation using only visual input~\citep{DBLP:conf/icra/SinopoliMDK01, DBLP:conf/icra/BloschWSS10} and natural language instruction based navigation~\citep{DBLP:conf/aaai/MacMahonSK06, DBLP:conf/acl/VogelJ10}. Furthermore, combining vision and language information during navigation task is more challenging and realistic, because natural language navigation instruction should be interpreted based on visual input, just like humans.

\subsection{Visual-Linguistic Learning}
The deep learning advances the development of both computer vision and natural language processing. Moreover, it unifies the visual and language data into a vector representation. This is the most important preliminary to combine vision with language and high-level reasoning. Visual Question Answering\citep{DBLP:conf/iccv/AntolALMBZP15,DBLP:conf/cvpr/JohnsonHMFZG17,DBLP:conf/cvpr/GoyalKSBP17,DBLP:conf/cvpr/HudsonM19} is a prime example of this trend. Others are image captioning~\citep{DBLP:journals/tacl/YoungLHH14,DBLP:journals/corr/ChenFLVGDZ15,DBLP:conf/acl/SoricutDSG18}, visual commonsense reasoning ~\citep{DBLP:conf/cvpr/ZellersBFC19,DBLP:journals/corr/abs-1908-02962} and so on.

\section{Datasets and Simulators}\label{sec:pf}
Considering of the scale, annotation and influence, we will introduce some typical datasets ans simulators. Other datasets or simulators will be briefly referred alone the introduction of related tasks.
\begin{table*}[!htb]
	\centering
	\caption{Statics of typical simulators involving VLN. $\dagger$ means that the dataset is  built in.}
	\begin{threeparttable}
		\begin{tabular}{cccc}
			
			Simulator & Dataset & Observation & 2D/3D \\ \toprule
			AI2-THOR$\dagger$ & - & photo-realistic  & 3D \\ 
			VizDoom$\dagger$ & - & Virtual  & 2.5D \\ 
			Gibson$\dagger$ & 2D-3D-S, Matterport3D & photo-realistic  & 3D \\ 
			LANI$\dagger$ & - & Virtual & 3D \\
			House3D & SUNCG & photo-realistic  & 3D \\ 
			Matterport3D Simulator & Matterport3D & photo-realistic & 3D \\ 
			Habitat & Gibson, Matterport3D, Replica & photo-realistic & 3D \\ 
			\bottomrule
		\end{tabular}
	\end{threeparttable}
	\label{tab:sim}
\end{table*}

\subsection{Datasets}

\textbf{SUNCG}~\citep{DBLP:conf/cvpr/SongYZCSF17}\footnote{\url{https://github.com/shurans/sscnet}} dataset contains 45,622 artificially designed 3D scenes, ranging from single chamber to multi-floor houses. These 3D scenes include a large number of objects, spatial layout and other elements, hoping to provide a good platform for 3D object recognition researchers.
On average, there are 8.9 rooms and 1.3 floors per scene, and also many categories of rooms and objects in each scene. More than 20 types of room and 80 categories of object in SUNCG datasets, 404,508 room instances and 5,697,217 object instances. 

People have annotated each scene in dataset with 3D coordinates and inside room and object types. At every time step an agent has access to the following signals: a) the visual RGB signal of its current first person view; b) semantic (instance) segmentation masks for all the objects visible in its current view; c) depth information. For different tasks, these signals might serve for different purposes, e.g., as a feature plane or an auxiliary target.  

\textbf{Matterport 3D}~\citep{DBLP:conf/3dim/ChangDFHNSSZZ17}\footnote{\url{https://github.com/niessner/Matterport}} is a recent mainstream dataset, which contains 90 scenes collected from reality and 194,400 RGB-D images in total. Dataset have been fully annotated with 3D reconstructions and 2D and 3D semantic segmentation. 

The precise global alignment of the entire building and a comprehensive and diverse panoramic view set supports various supervised and self-supervised computer vision tasks, such as view overlap prediction, semantic segmentation, and region classification etc.

\textbf{2D-3D-S}~\citep{DBLP:journals/corr/ArmeniSZS17}\footnote{\url{http://3Dsemantics.stanford.edu/}} is a indoor environment supporting 2D, 2.5D and 3D domains with multi-modal, 3D meshes and point clouds data. The dataset has over 70,000 RGB images, annotated with fully information such as depths,  semantic annotations,  and camera configurations.

\textbf{Replica}~\citep{DBLP:journals/corr/abs-1906-05797}\footnote{\url{https://github.com/facebookresearch/Replica-Dataset}} 
is a photo-realistic 3D indoor environment with 18 different scenes rendered by high quality images. 
The scenes selected focus on the semantic diversity and scale of the environment. Each scene is composed of dense grids, high-resolution textures, original semantic classes, instance information, flat mirrors and glass reflectors.

\subsection{Simulators with Builtin Dataset}

\textbf{LANI}~\citep{DBLP:conf/emnlp/MisraBBNSA18} is a massive 3D navigation dataset with 6,000 language instructions in total, based on Unity 3D. The environment is a fenced, square, grass field. Each scene includes between 6–13 randomly placed landmarks, sampled from 63 unique landmarks. The agent has discrete actions: forward, stop, turnleft, and turnright. At each time step, the agent performs an action, observes a first person view of the surroundings as an image, and receives a scalar reward. The simulator provides a socket API to control the agent and the environment.

\textbf{AI2-THOR}~\citep{DBLP:journals/corr/abs-1712-05474}\footnote{\url{http:// ai2thor.allenai.org.}}is a large-scale near photo-realistic 3D indoor dataset, where agents can navigate in the scenes and interact with objects to perform tasks. AI2-THOR enables research in many different domains, such as deep reinforcement learning, planning, visual question answering, and object detection and segmentation etc.

\textbf{VizDoom}\citep{DBLP:conf/cig/KempkaWRTJ16}\footnote{\url{http://vizdoom.cs.put.edu.pl}} is modified on Doom, a first-person shooter video game. Aiming at providing convenience for researchers, ViZDoom is designed small-scale, efficient, and highly customizable for different domains of experiments. Besides, VizDoom supports different control modes, custom scenes, access to the depth buffer, and 
can run without a graphical interface, which improving the efficiency of algorithm execution.

\textbf{Gibson}~\citep{DBLP:conf/cvpr/XiaZHSMS18}\footnote{\url{http://gibson.vision/}} is based on virtualized real spaces, with embodiment of agents and making them subject to constraints of complex semantic scenes. Gibson consists of 572 building scenes and 1,447 floors,  and each scene is equipped with panoramic images and camera configurations. The base format of the dataset is similar to 2D-3D-S dataset ~\citep{DBLP:journals/corr/ArmeniSZS17}, but is more diverse and has 2 orders of magnitude more spaces. The simulator of Gibson has also integrated 2D-3D-S dataset~\citep{DBLP:journals/corr/ArmeniSZS17} and Matterport3D~\citep{DBLP:conf/3dim/ChangDFHNSSZZ17} for optional use.

\subsection{Simulators}
\textbf{House3D}~\citep{DBLP:conf/iclr/WuWGT18}\footnote{\url{http://github.com/facebookresearch/House3D}} contains diverse room types and objects inside and support MacOS and Linux operating system. The 3D scenes are based on SUNCG dataset and annotated with rich information. To build a realistic 3D environment, House3D offers a OpenGL based renderer for the SUNCG scenes. Agents in this environment can move freely, receiving tasks for the different kinds of research.

\textbf{Matterport3D Simulator}~\citep{DBLP:conf/cvpr/AndersonWTB0S0G18}\footnote{\url{ https://github.com/peteanderson80/Matterport3DSimulator}} is a large-scale machine learning research platform based on the Matterport3D dataset~\citep{DBLP:conf/3dim/ChangDFHNSSZZ17} for the research and development of intelligent agents. Agents can virtually ‘move’ throughout a scene by adopting poses coinciding with panoramic viewpoints. Each scene has a corresponding weighted, undirected graph, so that the presence of an edge indicates a robot-navigable transition between the two viewpoints. The simulator does not define or place restrictions on the agent’s goal, reward function, or any additional context, so researchers can design this metrics according to their experimental settings.

\textbf{Habitat}~\citep{DBLP:conf/iccv/SavvaMPBKMZWJSL19}\footnote{\url{https://aihabitat.org}} is a platform for research in photo-realistic 3D environment, and integrates multiple commonly used VLN datasets. Specifically, the simulator supports Matterport3D~\citep{DBLP:conf/3dim/ChangDFHNSSZZ17}, Gibson~\citep{DBLP:conf/cvpr/XiaZHSMS18}, and Replica~\citep{DBLP:journals/corr/abs-1906-05797} datasets. Besides, Habitat consists of a simulator Habitat-Sim and a modular library Habitat-API.  Habitat-API aiming to help researchers verify and improve intelligent algorithms.

\section{Goal-Oriented Tasks}\label{sec:vis}

In goal-oriented tasks, the language instruction only contain specific goals with no detailed route on it, so agents need to make plan by themselves. In this section, we will introduce several goal oriented tasks from task definition, evaluation metrics, related works and typical methods.

\subsection{LANI}

\subsubsection{Task Definition}
\citet{DBLP:conf/emnlp/MisraBBNSA18} collect a corpus of navigation instructions using crowd sourcing. They used LANI simulator to generate environments randomly and one reference path for each environment. 
The generated reference paths are near to their neighbor landmarks to elicit instructions. Then, they used Amazon Mechanical Turk to annotate.
The experiment environment is simulated and relatively simple.  Furthermore, \citet{DBLP:conf/corl/BlukisTNKA19} propose a real-world learning framework, which is similar to LANI, so we will not describe details here.

\subsubsection{Evaluation Metrics}
The task performance is evaluated on a test set $\{(\bar{x}^{(i)}, s_1^{(i)}, s_g^{(i)} ) \}_{i=1}^M$, where $\bar{x}^{(i)}$ is an instruction, $s_1^{(i)}$ is the start state, and $s_g^{(i)}$ is the goal state. Task completion accuracy and the distance of the agent’s final state to $s_g^{(i)}$ are evaluated.

\subsubsection{Related Task}
Before LANI, \citet{DBLP:conf/rss/BlukisBBKA18} propose a virtual environment navigation task based on Unreal Engine, which uses the AirSim plugin \cite{DBLP:conf/fsr/ShahDLK17} to simulate realistic quadcopter dynamics. The agent is a quad-copter flying between landmarks. This work focus on the problem of mapping, planning and task execution. Language instructions for this task are generated from a pre-defined set of templates. For solving this problem, ~\citet{DBLP:conf/rss/BlukisBBKA18} put instructions into a LSTM network to get language embedding at the task beginning. A custom residual network is used to represent the image features at every time step and project features in global reference frame for making actions. 

\subsubsection{Typical Methods}

In~\citet{DBLP:conf/emnlp/MisraBBNSA18}'s work,  this instruction following task is decomposed into two submodules: goal prediction and action generation. They proposed a new language-conditioned image generation network architecture LINGUNET, to make a map from visual input to goals output.

Base on LANI, \citet{DBLP:conf/corl/BlukisMKA18} propose an approach for mapping natural language instructions and raw observations inputs to continuous control of a quadcopter drone, which used the quadcopter simulator environment from \citet{DBLP:conf/rss/BlukisBBKA18}. For indicating where the agent should visit during navigation and where to stop, they built a model to predict the position-visitation distributions, and then actions are generated from the predicted distributions.

For combing the simulation and reality, \citet{DBLP:conf/corl/BlukisTNKA19} propose a learning framework to map language instructions and images to low-level action outputs. In addition, Supervised Reinforcement Asynchronous Learning (SuReAL) is used in both simulation and reality without the need to fly in real world during training. SuReAL combines supervised learning for predicting next goal and reinforcement learning for continuous action outputs.

More recently, \citet{DBLP:journals/corr/abs-2011-07384} study the problem of extending to reason about new objects. Due to the lack of sufficient training data, they used a few-shot method trained from extra augmented reality data to ground language instructions to objects. This method can align the objects and their mentions in language instructions.

\subsection{Action Learning From Realistic Environments and Directives}

\subsubsection{Task Definition}

Action Learning From Realistic Environments and Directives~\cite{DBLP:conf/cvpr/ShridharTGBHMZF20}(ALFRED) is a indoor task, agents in this task need to receive a language instruction about doing a household and first-person image observations, then generated a sequence of actions to finish it. ALFRED includes 25,743 English instructions, describing 8,055 expert demonstrations, each with an average of 50 steps, resulting in 428,322 image-action pairs.The expert demonstrations are together with both high-level and low-level language instructions in 120 indoor scenes based on AI2-THOR 2.0 simulator~\cite{DBLP:journals/corr/abs-1712-05474}. These demonstrations involve partial observability, a long range of actions, in a designated natural language, and irreversible actions.

\subsubsection{Evaluation Metrics}
ALFRED allows users to evaluate both full task and task goal-condition completion. In navigation-only tasks, the model performance can be measured by how far the agent is from the goal. Whether the target conditions of the task have been completed can also be evaluated in the ALFRED task. The evaluation metrics consist of \textit{Task Success, Goal-Condition Success and Path Weighted Metrics}.

\begin{itemize}
	\item Task Success. If the object positions and state changes correspond correctly to the task goal-conditions at the end of the action sequence, the task success will be set to 1, and otherwise to 0.
	\item Goal-Condition Success. This metric is the ratio of goal-conditions completed at the end of an episode to those necessary to have finished a task.
	\item Path Weighted Metrics. \citet{DBLP:journals/corr/abs-1807-06757} find that PDDL solver can not find the optimal result, but usually efficient. The path weighted score $p_s$ for metric $s$ is given as
	\begin{equation}
		p_s = s \times \frac{L^*}{max(L^*, \hat{L})}
	\end{equation}
	where $\hat{L}$ is the number of actions the model took in the episode, and $L^*$ is the number of actions in the expert demonstration. Obviously, a model receives half-credit for taking twice as long as the expert to accomplish a task.
\end{itemize}

\subsubsection{Typical Methods}

In~\cite{DBLP:conf/cvpr/ShridharTGBHMZF20}, a Sequence-to-Sequence (Seq2Seq) Model with progress monitoring is introduced. Each visual observation was encoded with a froze ResNet-18~\cite{DBLP:conf/cvpr/HeZRS16}. The step-by-step instructions is combined into a single input sequence with the $<SEP>$ token. Then the sequences are fed into a bi-directional LSTM encoder to produce an encoding sequence. The agent's actions at each time step are based on the attention mechanism that can identify the related tokens in the language instruction. The agent interacts with the environment by choosing an action and producing dense pixel-wise binary mask to indicate specific objects in the frame.

In~\cite{DBLP:journals/corr/abs-2101-03431}, during the navigation of the sub-goals, the agent's field of view is enhanced through multiple perspectives, and the agent is trained to predict its relative spatial relationship with the target position at each time step.

\citet{DBLP:journals/corr/abs-2012-03208} propose a \textit{Modular Object-Centric Approach} (MOCA) to decouple ALFRED task into two sub-task, i.e., visual perception and action policy generation. The former aiming to predict an interaction mask for the object that the agent interacts with using object-centric mask prediction, while the latter makes a prediction of the current action of the agent.

\subsection{Embodied Question Answering}

\subsubsection{Task Overview}

Embodied Question Answering (EQA) tasks require an agent to ask a question (e.g. `What color is the cup?'). The agent usually situated at a random location in an environment (a house or building), and can observe the environment by first-person perspective images. The action space of the agent includes move forward, turn and strafe, etc. After understanding the question, the agent needs to gather useful information in the environment for answering it. Unlike the previous work~\cite{yu2017deep}, the agent does not receive any global or structured representation of the environment (map, location, rooms, objects), or of the task (the functional program that generated the question).

EQA is built upon House3D simulator and SUNCG datasets, more details about House3D please refer to section~\ref{sec:pf}. In order to ensure the quality of the dataset, the environment inside must be realistic and typical, and there must be no abnormal conditions. 
Scenes in this task have one kitchen, dining room, living room, and bedroom at least. The EQA v1 dataset contains more than 5000 questions in over 750 environments, involving a total of 45 unique objects in 7 unique room types. There are 1 to 22 related questions in each environment, with an average of 6. Preposition questions are less than other kinds of questions because many frequently occurring spatial relationships are too easy to solve without exploration and cannot be processed by entropy threshold. 

\subsubsection{Evaluation Metrics}

The goal of an EQA agent is to answer questions precisely. However, it is important to disentangle success/failure at the intermediate task of navigation from the downstream task of question answering. The evaluation metrics of this task are as following:
\begin{itemize}
	\item Question Answering Accuracy. The mean rank of the ground-truth answer in the candidate answers list sorted by the model.
	\item Navigation Accuracy. This metric consists of the distance to the target object at navigation termination $\text{d}_\text{T}$, changes in distance to target from initial to final position $\text{d}_{\Delta}$, and the smallest distance to the target at any point in the episode $\text{d}_\text{min}$. All these distances are measured in meters along the shortest path to the target.
	\item The percentage of questions for which an agent either terminates in ($ \% \text{r}_\text{T}$) or ever enters ($ \% \text{r}_{\hookleftarrow}$) the room containing the target objects.
	\item The percentage of episodes in which agents choose to stop navigation and give a answer before reaching the maximum episode length ($\% \text{stop}$).

\end{itemize}

\subsubsection{Typical Methods}

\citet{DBLP:conf/cvpr/DasDGLPB18} give a baseline model, which has four modules, i.e., vision, language, navigation, answering and are trained in two stages. First, the navigation and answer module use imitation/supervised learning to independently train the automatically generated navigation expert demonstrations. Then, the navigation architecture uses policy gradients for fine-tuning. This model can map raw sensory input to navigation actions and answers generation.

\citet{DBLP:journals/corr/abs-1811-05013} explore blindfold (question-only) model for EQA which neglects environmental and visual information. Intuitively, the blindfold model is a degraded solution, but it reaches state-of-the-art results on EQA except some rare initialization situations.

\citet{DBLP:conf/corl/DasGLPB18} propose a modular approach of learning strategies for navigating within long-term planning from instructions. For increasing sample efficiency, imitation learning is used to warm-start policies at each level of the hierarchical policy on multiple timescales.

\citet{DBLP:conf/nips/ParvanehATSH20} propose a new learning strategy that learns both from observations and generated counterfactual environments. Specifically, the algorithm generating counterfactual observations on the fly for navigation is introduced as the linear combinations of present environments. In addition, the agent’s actions are encouraged to remain stable between initial and counterfactual environments by training objective–effectively removing false features that could mislead the agent.

Furthermore, \citet{DBLP:journals/tip/WuJY20} introduce an easier and practical EQA setting called \textit{calibration}. They designed a warm-up stage that the agent is asked a few rhetorical questions when it entered into a new environment. The goal is to adapt the agent policy to the new environments. Then they proposed a new model, which contains two modules, designed for the intermediate navigation task, called Navigation Module, and downstream question answering task, called Question Answering Module, respectively.

\subsubsection{Task Variation}
When using behavioral cloning method to train a recurrent model for navigation, the impact of a novel loss-weighting metric named inflection weighting has grown. 
\citet{DBLP:conf/cvpr/WijmansDMDGLEPB19} extend the EQA task in realistic environments from the Matterport3D dataset~\citep{DBLP:conf/3dim/ChangDFHNSSZZ17} and propose a model which can surpass the baseline by optimize the infection weighting metric. Moreover, they argue that point clouds can provide more information than traditional RGB images in photo-realistic environment, especially for tasks like embodied navigation and obstacle avoidance.

\citet{DBLP:conf/cvpr/YuCGBBB19} extend the EQA to Multi-Target version, \ie, Multi-Target EQA. As the questions are more complex, the agent is supposed to navigate to multiple locations and performs some comparative reasoning steps before answering the question. A new model is designed to solve MT-EQA, which consists of four modules: the question-to-program generator, the navigator, the controller, and the VQA module.

\subsection{RoomNav}

\subsubsection{Task Overview}
While \citet{DBLP:conf/iclr/WuWGT18} propose the House3D simulator, a Concept-Driven Navigation benchmark task is also proposed (\ie, RoomNav). The goal is defined as "Go to X", where X represents a pre-defined room type or object type. This is a semantic concept, and the agent needs to be interpreted from a variety of scenes with different visual appearances. RoomNav consists of 270 houses divided into three splits, \ie, small, large and test with 20, 200, and 50 respectively.

\textbf{Observation:} Three different kinds of visual input signals are utilized for $X_t$, including (1) raw pixel values; (2) semantic segmentation mask of the pixel input;  (3) depth information, and different combinations of these three. Each concept (language instruction) $I$ is encoded as a one-hot vector representation.

\textbf{Action Space:} This task support continuous and discrete actions, such as agents can move in certain velocities or pre-defined movements.

\subsubsection{Evaluation Metrics}
The episode is regarded as successful if both of the following two criteria are satisfied: (1) the agent is located inside the target room; (2) the agent consecutively sees a designated object category associated with that target room type for at least 2 time steps. An agent sees an object if there are at least 4\% of pixels in $X_t$ belonging to that object.

\subsubsection{Typical Methods}

In~\citep{DBLP:conf/iclr/WuWGT18}, a gated-attention architecture is introduced. Gated-CNN and Gated-LSTM network are used for controlling continuous and discrete actions respectively. The Gated-CNN policy is trained using the Deep Deterministic Policy Gradient (DDPG)~\citep{DBLP:journals/corr/LillicrapHPHETS15}, while the Gated-LSTM policy is trained using the asynchronous advantage actor-critic algorithm (A3C)~\citep{DBLP:conf/icml/MnihBMGLHSK16}.

\citet{DBLP:conf/iccv/WuWTRGT19} introduce a new memory architecture, \textit{Bayesian Relational Memory} (BRM), taking the form of a probability relationship graph on semantic entities to improve the generalization ability in unseen environments. 
BRM supports obtaining some prior knowledge from the training environment, and at the same time, it can update the knowledge in memory after exploration. A BRM agent consists of a module for generating sub-goals and goal-conditioned locomotion module for control

\subsection{Remote Embodied Visual referring Expression in Real Indoor Environments}
\subsubsection{Task Overview}
Remote Embodied Visual referring Expression in Real Indoor Environments - REVERIE~\citep{DBLP:conf/cvpr/QiW0WWSH20} 
task requires the intelligent agent to correctly locate the remote target object specified by the concise high-level natural language instructions (which cannot be observed at the starting position). Since the target object and the starting object are in different rooms, the agent needs to navigate to the target location first. Unlike other navigation tasks, agents in this task can execute detection action if agents thinks they has localised the target object and decide to output the region of it. After the detection action, this episode is finished.

\subsubsection{Evaluation Metrics}
The performance of a model is mainly measured by REVERIE success rate, which is the number of successful tasks over the total number of tasks. A task is considered successful if it selects the correct bounding box of the target object from a set of candidates. Since the target object can be observed at different viewpoints or camera views, an episode is successful as long as the agent can identify the target within 3 meters, regardless of from different viewpoints or views. We also measure the navigation performance with four kinds of metrics, including success rate, oracle success rate, success rate weighted by path length, and path length (in meters)~\citep{DBLP:conf/cvpr/AndersonWTB0S0G18}. 

\subsubsection{Typical Methods}
~\citet{DBLP:conf/cvpr/QiW0WWSH20} propose a modular model including three parts: a navigator module that decides the action to take for the next step, a pointer module that attempts to localise the target object according to the language guidance, and an interaction module that responses for sending the referring expression comprehension information obtained from the pointer to the navigator to guild it to make more accurate action prediction.

\citet{DBLP:journals/corr/abs-2103-12944} introduce two pre-training tasks, called Scene Grounding task and Object Grounding task, and a new memory-augmented attentive action decoder. The pre-training tasks encourage agents learn where to stop and what to attend to, and the action decoder uses past observations to merge visual and textual information in an effective way.

\subsection{Other Tasks}

\subsubsection{Talk2Car}

Talk to car task is proposed by~\citet{DBLP:conf/emnlp/DeruyttereVGGM19}, and the dataset for this task is built upon the \textit{nuScenes} dataset~\cite{DBLP:conf/cvpr/CaesarBLVLXKPBB20}. This dataset includes 11,959 instructions for the 850 videos of nuScenes training set since 3D bounding box annotations for the test set of nuScenes  are not available. 55.94\% and 44.06\% of these commands belong to videos taken respectively in Boston and Singapore. On average a command consist of 11.01 words, 2.32 nouns, 2.29 verbs and 0.62 adjectives. Each video has on average 14.07 commands. The textual annotations are very complete, it not only indicates what to do, such as `pick him up', but how to drive the car, such as `turn around'. However, this task did not give a metric to evaluate the performance of the action. 

\citet{DBLP:conf/emnlp/DeruyttereVGGM19} only focus on detecting the referred object, but ignoring the action execution. They only assessed the performance of 7 detection models to find the objects in a command. In the test set, random noun matching approach is used to match the nouns in command with the visual objects.

\subsubsection{XWORLD}

This task is proposed by~\citet{DBLP:conf/iclr/YuZX18}. It consists of XWORLD2D and XWORLD3D.
\begin{itemize}
	\item XWORLD2D. To test the method of interactive grounded language acquisition and generalization, ~\citet{DBLP:conf/corl/YuLZX18} build a 2D maze-like world, \ie, XWORLD2D, 
	including language related navigation task and question answering task. Specifically, the navigation task requires agent to follow the language instructions to navigate the final destination, and question answering task requires agents to generate one-word answer for questions.
	\item XWORLD3D. The XWORLD3D is the extension of XWORLD2D where changing the environment from full-observability to partial-observability, and adding two actions \textit{Turn left} and \textit{Turn right} to the discrete action set. For increasing the visual variance, at each session the authors randomly rotate each object and scale it randomly within $[0.5, 1.0]$. Suppose each map is $X \times Y$ , the maximum time steps of an episode is $3 X Y$.
\end{itemize}

In addition, successful episode means the agent acts obey the teacher’s command. The failure episode is triggered whenever the agent hits any object that is not required by the command. The success rate is used to evaluation the model performance.

In~\cite{DBLP:conf/corl/YuLZX18}, an effective baseline method \textit{Guided Feature Transformation} (GFT) is proposed for language grounding. In order to improve the fusion of visual and language features, the latent sentence embeddings calculated from the language input is regarded as the conversion matrix of visual features.
This method is completely differentiable and embedded in the agent's perception system, which is trained end-to-end by the RL. GFT method can adapt to both 2D and 3D environments without changing any architecture or hyperparameters between the two scenarios.

\subsubsection{3D Doom}
\citet{DBLP:conf/aaai/ChaplotSPRS18} create an environment for language grounding research, where the agent needs to follow natural language instructions and receives positive rewards after successfully completing the task. This environment is built on top of the VizDoom API~\cite{DBLP:conf/cig/KempkaWRTJ16}, based on Doom, a classic first person shooting game. 

In 3D Doom task, the instruction is a (action, attribute, object) triplet. Each instruction can have multiple attributes, but the number of actions and objects is limited to one each. This environment allows various objects to be generated at different locations on the map. Objects can have various visual attributes. Success rate defined as the proportion of success navigation times in the total number of navigation times. There are two scenarios for evaluation:

\begin{itemize}
	\item Multi-task Generalization. The agent uses the instructions in the training set for evaluation on an unseen map. The unseen map consists of a combination of invisible objects placed at random locations. 
	\item Zero-shot Task Generalization, where the agent is situated in unseen environment with new combinations of attribute-object pairs which are not seen during the training. The maps in this scenario are also unseen.
\end{itemize}

The model proposed by~\cite{DBLP:conf/aaai/ChaplotSPRS18} uses the Gated-Attention mechanism to combine images and text representations, and utilize standard reinforcement and imitation learning methods to learn strategies for executing natural language instructions. This model is end-to-end and consists of State Processing Module and Policy Learning Module.
\subsubsection{Visual Semantic Navigation}

This task~\cite{DBLP:conf/iclr/YangWFGM19} is based on the interactive environments of AI2-THOR~\cite{DBLP:journals/corr/abs-1712-05474}. 
There are 87 object categories within AI2-THOR that are common among the scenes. However, some of the objects are not visible without interaction. Therefore, only 53 categories based on their visibility at random initialization of the scenes. To test the generalization ability of the method on novel objects, these 53 object categories are split into known and novel sets, and the known set of object categories are used in training. Only the navigation commands of AI2-THOR are used for the task. These actions include: move forward, move back, rotate right, rotate left, and stop.

The evaluation metrics are Success Rate and the Success weighted by Path
Length (SPL) metric recently proposed by~\citet{DBLP:journals/corr/abs-1807-06757}.
\begin{itemize}
	\item Success Rate is defined as the ratio of the number of times the agent successfully navigates to the target and the total number of episodes. 
	\item SPL is a better metric which is a function considering both Success Rate and the path length to reach the goal from the starting point. It is defined as $\frac{1}{N} \sum_{i=1}^N S_i \frac{L_i}{max(P_i, L_i)}$, where $N$ is the number of episodes, $S_i$ is a binary indicator of success in episode $i$, $P_i$ represents path length and $L_i$ is the shortest path distance (provided by the environment for evaluation) in episode $i$.
\end{itemize}

Based on the actor-critic model~\cite{DBLP:conf/icml/MnihBMGLHSK16}, \citet{DBLP:conf/iclr/YangWFGM19} propose to use \textit{Graph Convolutional Networks} (GCNs)~\cite{DBLP:conf/iclr/KipfW17} to incorporate the prior knowledge into a Deep Reinforcement Learning framework. The knowledge of the agents is encoded in a graph. GCNs can encode various structured graphs in an efficient way.

\subsubsection{Behavioral Robot Navigation}

\citet{DBLP:conf/emnlp/ZangPVCNSS18} create a new dataset for the problem of following navigation instructions under the behavioral navigation framework of~\citep{DBLP:conf/icra/SepulvedaNS18}. This dataset contains 100 maps of simulated indoor environments, each with 6 to 65 rooms. It consists of 8066 pairs of free-form natural language instructions and navigation plans for training. This training data is collected from 88 unique simulated environments, 6064 distinct navigation plans in total (2002 plans have two different navigation instructions each, and the rest has one). The dataset contains two different testsets:
\begin{itemize}
	\item Test-Repeated: uses seen trainset environments and unseen routes, including 1012 pairs of instructions and navigation plans.
	\item Test-New: uses 12 unseen environments and unseen routes, which is more complex, including 962 pairs of instructions and navigation plans.
\end{itemize}

An end-to-end deep learning model is proposed by~\citet{DBLP:conf/emnlp/ZangPVCNSS18} to convert free-form natural language instructions into high-level navigation plans.

\section{Route-Oriented Tasks}\label{sec:nat}

In route-oriented tasks, the language instruction describes the objects seen along the way and the agent's route in detail. If the agent can understand the instructions and navigate according to the path in the instructions, it can reach the final destination and successfully complete the navigation task. In this section, we will introduce several goal oriented tasks from task definition, evaluation metrics, related works and typical methods.

\subsection{Room-to-Room}

\subsubsection{Task Description}
Room-to-Room (R2R) is a natural language navigation dataset based on vision in a photo-realistic environment~\cite{DBLP:conf/cvpr/AndersonWTB0S0G18}. Environments in this dataset is defined by a navigation graph, where nodes are locations with a self-centered panoramic image, and edges define effective connections for agents navigation. This dataset(spited with train/validation/test)  is fully annotated with instructions and paths and each path has 3 different related instructions. The average length of ground-truth path is 4 to 6 edges. 

R2R is a task where the inputs are the images from egocentric panoramic camera and instruction. Note that images are updating with the motion of agent, while instruction is given at the beginning. For example, given an instruction \textit{``Go towards the front door but before the front door make a left, then through the archway, go to the middle of the middle room and stop."}, an agent is supposed to understand the instruction and follow it from start point to the goal position as soon as possible. 
\subsubsection{Evaluation Metrics}
There are a bunch of metrics to evaluate a R2R agent. Typical used metrics are Success Rate, Navigation Error, Path Length and Success weighted by Path Length. The Oracle Navigation Error, Oracle Success Rate can be used to monitor the training process.
\begin{itemize}
	\item Success Rate (SR) shows how many times the last node of the predicted path is within a certain threshold distance of the last reference path node.
	\item Navigation Error (NE) measures the distance between the last predicted path node and the last reference path node.
	\item Path Length (PL) is equal to the total length of the predicted path. It is optimal when it’s equal to the length of the reference path.
	\item Success weighted by Path Length (SPL) \cite{DBLP:conf/cvpr/AndersonWTB0S0G18} takes both success rate and path length into account and can be calculated in the same way as described in the previous section. A drawback of SPL is that it does not consider the similarity of the intermediate nodes of the predicted path to the reference path. This results in that although SPL score is high, the predicted path does not actually follow the  instructions, but just finds the correct target.
	\item Oracle Navigation Error (ONE) measures the smallest distance from any node in the path to the reference goal node.
	\item Oracle Success Rate (OSR) measures how often any node in the path is within a certain threshold distance from the goal. The goal is represented by the last node of the reference path.
\end{itemize}

Due to the inherent limitations of the existing evaluation metrics, the performance of the model cannot be evaluated objectively and comprehensively. There are still many subsequent works to propose some new metrics, which are also introduced here.

\citet{huang2019multi} propose a discriminator model that can predict how well a given instruction explains the paired path, showing that only a small portion of the augmented data in \cite{fried2018speaker} are high fidelity. This metric can evaluate the instruction-path pair, which improves training efficiency and reduces the training time cost.
\citet{ilharco2019general} introduce the \textit{normalized Dynamic Time Warping} (nDTW) metric. nDTW slightly penalizes deviations from the ground-truth path and it is naturally sensitive to the order of the nodes composing each path. Besides, nDTW is suited for both continuous and graph-based evaluations, and can be efficiently calculated.
Despite the performance of R2R model has improved rapidly, current research is not clear about the role language understanding in this task, since mainstream evaluation metrics are focused on the completion of the goal, rather than the sequence of actions corresponding to the language instructions. 
Therefore,~\citet{DBLP:conf/acl/JainMKVIB19} analyze the drawbacks of current metrics for the R2R dataset and designed a novel metric named \textit{Coverage weighted by Length Score} (CLS). CLS measures how closely an agent’s trajectory fits with ground-truth path, not just the completion of the goal.
To improve the ability to evaluate language instructions, \citet{zhao2021evaluation} propose an instruction-trajectory compatibility model that operates without reference instructions. For system-level evaluations with reference instructions, SPICE metric~\cite{anderson2016spice} is better than BLEU~\cite{papineni2002bleu}, ROUGE~\cite{lin2004rouge}, METOR~\cite{denkowski2014meteor} and CIDEr~\cite{vedantam2015cider}. In other situations (e.g., selecting individual instructions, or model selection without reference instructions) using a learned instruction-trajectory compatibility model is recommended.

\subsubsection{Typical Methods}
The R2R task is most impressive, and most research works of VLN are based on it. Basic framework of these works are Seq2Seq model, but they focus on different aspects to improve the model performance, therefore we divide them to \textbf{7 classes} for clear understanding, i.e., \textit{Exploration Strategy, Beyond Seq2Seq Architecture, Reinforcement Learning and Imitation Learning, Language Grounding, Data Augmentation, Pre-Traning model and other related research.}

\textbf{Exploration Strategy.} 
In this class, related works focus on finding a effective and fast path from start position to destination.~\citet{ma2019self} proposes a module to estimate the progress that made by agent towards the goal. 
Based on that, \citet{ma2019regretful} design two modules for exploration, \ie, Regret Module for moving forward or rolling back and Progress Marker module to help the agent decide which direction to go next by showing directions that are visited and their associated progress estimate.
While all current approaches make local action decisions or score entire trajectories using beam search,~\citet{ke2019tactical} present the \textit{Frontier Aware Search with backTracking} (FAST) navigator. When the agent realizes lost itself, FAST navigator can explicit backtrack by using asynchronous search. This can also be plug and play on other models.
\citet{huang2019transferable} define two in-domain sub-tasks: \textit{Cross-Modal Alignment} (CMA) and \textit{Next Visual Scene} (NVS). The visual and textual representations of the agent learned in certain environment can be transferred to other environments with the help of CMA and NVS.
\citet{zhu2020vision} introduce four self-supervised auxiliary reasoning tasks to take advantage of the additional training signals derived from the semantic information. These auxiliary tasks improve the model performance by giving more semantic information to help the agent to understand environments and tasks.
\citet{wang2020active} introduce an end-to-end architecture for learning an exploration policy, which enables the agent intelligently interacts with the environment and actively gather information when faced with ambiguous instructions or unconfident navigation decisions.
In~\cite{zhang2020language}, a learning paradigm using recursive alternate imitation and exploration is proposed to narrow the difference between training and testing stages. 
\citet{deng2020evolving} introduce the \textit{Evolving Graphical Planner} (EGP), a method that uses raw images to generate global navigation plan efficiently. It has the advantage of dynamically building a graphical representation and generalizing the action space to allow more flexible decision-making.

\textbf{Beyond Seq2Seq Architecture.}
Encouraged by recent progress on attention networks, many transformer-based studies have emerge, and we classify them as Beyond Seq2Seq architecture.

\citet{landi2019perceive} devise \textit{Perceive, Transform, and Act} (PTA) architecture to use the full history of previous actions for 
different modalities.
\citet{magassouba2021crossmap} propose the \textit{Cross-modal Masked Path } transformer, which encodes linguistic and environment state features to sequentially generate actions similarly to recurrent network approaches. In addition, this method uses feature masking to better model the relationship between the instruction and environment features.
\citet{wu2021improved} and \citet{mao2020vision} use multi-head attention on visual and textual input to enhance the performance of the model.
To capture and utilize the inter-modal and intra-modal relationships among different scenes, its objects, and directional clues, \citet{hong2020language} devise language and visual entity relationship graph model. A message-passing algorithm is also proposed to spread information between the language elements and visual entities in the graph, and then the information are combined by the agent to determine the next action.
\citet{xia2020multi} present a novel training paradigm, \textit{Learn from Everyone}(LEO), which utilizes multiple language instructions (as multiple views) for the same path to resolve natural language ambiguity and improve the model generalization capabilities.~\citet{qi2020object} distinguish the object and action information from language instructions while most existing methods pay few attentions, and propose a \textit{Object-and-Action Aware Model} (OAAM) that processes these two different forms of natural language based instruction separately. OAAM enables each process to flexibly match object-centric (action-centric) instructions with their corresponding visual perception (action direction).
To capture environment layouts and make long-term planning, \citet{wang2021structured} present \textit{Structured Scene Memory} (SSM), which allows the agent to access to its past perception and explores environment layouts. With this expressive and persistent space representation, the agent shows advantages in fine-grained instruction grounding, long-term reasoning, and global decision-making problem.

\textbf{Reinforcement Learning and Imitation Learning.}
Many works have found it is beneficial to combine imitation learning and reinforcement learning with VLN models. All of \citet{wang2019reinforced}, \citet{tan2019learning},  \citet{hong2020language}, \citet{parvaneh2020counterfactual}  and \citet{wang2020active} trained their models with two distinct learning paradigms, \ie, 1) imitation learning, where the agent is forced to mimic the behavior of its teacher. 2) reinforcement learning, which can help the agent explore the state-action space outside the demonstration path.
Specifically, in imitation learning, the agent takes the teacher action $a_{t}^{*}$ at each time step to efficiently learn to follow the ground-truth trajectory. In reinforcement learning, the agent samples an action $a_{t}^{s}$ from the action probability $p_{t}$ and learns from the rewards, which allows the agent to explore the environment and improves ability of generalization. Combining IL and RL balances exploitation and exploration when learning to navigate, formally, the following loss function can be used to optimize the model :
\begin{equation}
 \mathcal{L} = \lambda \underbrace{\sum_{t=1}^{T}-a_{t}^{*}\log(p_{t})}_{\mathcal{L}_{IL}}+\underbrace{\sum_{t=1}^{T}-a_{t}^{s}\log(p_{t})A_{t}}_{\mathcal{L}_{RL}}  
\end{equation}

For improving generalization ability, \citet{wang2018look} integrate model-based and model-free reinforcement learning methods which is a hybrid planned-ahead model and outperforms than baselines.
~\citet{wang2019reinforced} use \textit{Reinforced Cross-Modal Matching} (RCM) method with extrinsic and intrinsic rewards. The novel cycle-reconstruction reward is introduced to match the instruction and trajectory globally.
In~\cite{lansing2019valan}, based on SEED RL architecture, \textit{Vision And Language Agent Navigation}(VALAN) is introduced as a lightweight and extensible framework for learning algorithm research.
In order to reduce the impact of reward engineering design, \textit{Soft Expert Reward Learning} (SERL) model is proposed by~\citet{wang2020soft}, which contains two auxiliary modules: Soft Expert Distillation and Self Perceiving. The former activates agent to explore like an expert while the latter enforces the agent to reach the goal as soon as possible.
\citet{zhou2020inverse} devise a adversarial inverse reinforcement learning method to learn a language-conditioned policy and reward function. For better adapt to the new environment, variational goal generator is used during training to relabel the trajectory and sample different targets.

\textbf{Language Grounding.}
This class of method wants to ground language instructions better through the method of multimodal information fusion, so as to improve the success rate of agent navigation.
\citet{wang2019reinforced} present a novel cross-modal grounding architecture to ground language on both local visual information and global visual trajectory.
\citet{hu2019you} propose to decompose the grounding procedure into a set of expert models with access to different modalities (including object detection) and ensemble them at prediction time for utilizing all the available modalities more efficiently. 
For

For making the agent has a better understanding of the correspondence between text and visual modalities, a cross-modal grounding module composed of two complementary attention mechanisms is designed by~\citet{zhang2020language}.
\citet{kurita2020generative} build a neural network to compute the probability distribution over all possible instructions, and use Bayes’ rule to build a language-grounded policy. This method has a better interpretability than the traditional discriminative method by conducting comprehensive experiments.
\citet{hong2020sub} argue that the granularity of the navigation task should be at the level of these sub-instructions, rather than attempting to ground a specific part of the original long and complex instructions without any direct supervision or measuring navigation progress at word level.

\textbf{Data Augmentation.}
Using automatically generated navigation instructions as additional training data can enhance the model performance.  ~\citet{fried2018speaker}propose a speaker-follower framework for data augmentation and reasoning in supervised learning.
\citet{hong2020sub} enrich the benchmark dataset R2R with sub-instructions and their corresponding paths, \ie,  \textit{Fine-Grained Room-to-Room} (FGR2R). By pairing the sub-commands with the corresponding viewpoints in the path, FGR2R can better give the agent sufficient semantic information during the training process.
\citet{agarwal2019visual} present a work-in-progress ``speaker" model that generates navigation instructions in two steps. 
First, hard attention is used to select a series of discrete visual landmarks along the trajectory, and then a language conditioned on these landmarks is generated. 
Different from traditional data augmentation methods,  \citet{parvaneh2020counterfactual} propose an efficient algorithm to generate counterfactual instances that do not depend on hand-made or the particular field of rules. These counterfactual instances are added to the training, 
improving the ability of the agent when is tested in new environments.
In~\cite{fu2020counterfactual}, counterfactual idea has another application. A model agnostic method called \textit{Adversarial Path Sampler} (APS) is introduced to sample paths to optimize the agent navigation strategies gradually.
\citet{yu2020take} simultaneously deal with the scarcity of data in the R2R task while removing biases in the dataset through random walk data augmentation. By doing so, they are able to reduce the generalization gap and outperform baselines in navigating unknown environments.
\citet{DBLP:journals/corr/abs-2107-07201} designe a module named \textit{Neighbor-View Enhanced Model} (NvEM) to adaptively fuse the visual context from the neighbor views at the global level and the local level.
\citet{DBLP:journals/corr/abs-2106-07876} propose \textit{Random Environmental Mixup} (REM) aiming to reduce the performance gap between the seen and unseen environment and improve the overall performance. This method breaks up the environment and the corresponding path, and then recombines them according to certain rules to construct a brand-new environments as training data.
\citet{sun2021depth} point out that depth as a valuable signal source for the navigation has not yet fully explored and thus been ignored in previous research. Hence, they propose a \textit{Depth-guided Adaptive Instance Normalization module and Shift Attention} (DASA) module to address this issue. 

\textbf{Pre-Traning model.}
The general feature representation obtained through pre-training model can be applied to various tasks, and has been confirmed in many fields. A strong pre-trained backbone network can be effective in downstream task, like image recognition in CV and question answering in NLP. In particular, Transformer networks~\citet{DBLP:conf/nips/VaswaniSPUJGKP17} pre-trained with “masked language model”  objective~\citep{DBLP:conf/naacl/DevlinCLT19} on large language corpus outperforms on majority NLP tasks. Furthermore, \citet{DBLP:conf/iclr/SuZCLLWD20} developed VL-BERT, a pre-trainable generic representation for visual-linguistic tasks. In VLN field, pre-training approaches are used in two aspects:

\begin{itemize}
	\item \textbf{Using pre-trained models to solve VLN tasks. } In~\citet{DBLP:conf/emnlp/LiLXBCGSC19}, the agent is trained with pre-trained language models, (\ie, BERT~\citep{DBLP:conf/nips/VaswaniSPUJGKP17} and GPT-3~\citep{DBLP:conf/nips/BrownMRSKDNSSAA20}) and uses stochastic sampling to generalize well in the unseen environment. The \textit{PRE-trained Vision And Language based Navigator} (PREVALENT) model proposed by~\citet{DBLP:conf/cvpr/HaoLLCG20} is pre-trained with image-language-action triples, and fine-tuned on the R2R task. Based on the use of the PREVALENT model, the agent can better complete the task in an unseen environment.
	Since the pre-training model usually has massive parameters, the efficiency of using it will be relatively low. For reducing the scale of pre-train model and improving the efficiency of inference ,~\citet{DBLP:conf/ithings/HuangHZMLZS20} introduce two lightweight methods, \ie, factorization and parameter sharing
	based on the PREVALENT\citep{DBLP:conf/cvpr/HaoLLCG20} model. 
	\item \textbf{Pre-trained VL or VLN models on VLN tasks.} Due to the limitation of training data in VLN task,~\citet{majumdar2020improving} try to use massive web-scraped resources to address this issue. Therefore, VLN-BERT is proposed and proved that pre-training it on image-text pairs from the web before fine-tuning the specific path instruction data significantly improves the performance of VLN task.
	\citet{hong2020recurrent} propose VLN$\circlearrowright$BERT,
	which is a multi-modal BERT model equipped with a time-aware recurrent function to provide the agent with richer information.
	\citet{DBLP:journals/corr/abs-2104-04167} propose \textit{Object-and-Room Informed Sequential BERT} (ORIST) to improve the language grounding performance by encoding visual and instruction inputs at the same fine-grained level,  \ie, objects and words. Besides, the trained model can recognize the relative direction of each navigable location and the room type of its current and final navigation target.
\end{itemize}

\textbf{Other Related Research.}
To assess the implications of this work for robotics, \citet{anderson2020sim} transfer a VLN agent trained in simulation to a physical robot. There is a big difference between high-level discrete action space learned by the agent and robot’s low-level continuous action space. In order to address this issue, they introduce a sub-goal model to identify nearby navigable points and use domain randomization to reduce visual domain differences.
With the slow-down performance improvements in VLN tasks, 
a series of diagnostic experiments are carried out in ~\cite{zhu2021diagnosing} to reveal the agent's key points in the navigation process. The results show that the indoor navigation agent will refer to the object mark and direction mark in the instruction when making a decision. When it comes to visual and verbal alignment, many models claim that they can align object marks with certain visual targets, but there are still doubts about the reliability of this alignment.

\subsubsection{Model Performance Comparison}
A table comparing all models in R2R task. The state-of-the-art method is REM, since it is the effective combination of pre-trained backbone model (VLN$\circlearrowright$BERT) and data augmentation, making REM model the best existing approach on VLN benchmark.
\begin{table*}[t]
	\small
	\begin{center}
		\resizebox{1.0\textwidth}{!}{
			\setlength{\tabcolsep}{1.0em}
			{\renewcommand{\arraystretch}{1.0}
				\begin{tabular}{|l | c c  c c | c c c c | c  c c | }
					\hline
					Leader-Board (Test Unseen) 
					& \multicolumn{4}{c|}{Single Run} 
					& \multicolumn{4}{c|}{Pre-explore}
					& \multicolumn{3}{c|}{Beam Search}\\
					\hline
					Models
					& NE$\downarrow$ & OR$\uparrow$ & SR$\uparrow$& \textbf{SPL}$\uparrow$
					& NE$\downarrow$ & OR$\uparrow$& SR$\uparrow$ & \textbf{SPL}$\uparrow$
					& TL$\downarrow$ & SR$\uparrow$ & \textbf{SPL}$\uparrow$\\
					\hline
					Random~\cite{DBLP:conf/cvpr/AndersonWTB0S0G18} &  9.79 &  0.18 &  0.17 & 0.12  & - & - & - & - & - & - & - \\
					Seq-to-Seq~\cite{DBLP:conf/cvpr/AndersonWTB0S0G18} &  20.4 &  0.27 &  0.20 & 0.18  & - & - & - & - & - & - & -\\
					\hline
					Look Before You Leap~\cite{wang2018look} &  7.5 &   0.32 & 0.25  & 0.23  &-   &-  &- &- & - & - & -\\
					Speaker-Follower~\cite{fried2018speaker} &  6.62 &   0.44 & 0.35  & 0.28  &-  &-  &- &- &1257 &0.54 &0.01\\
					Chasing Ghosts~\cite{anderson2019chasing} & 7.83 &  0.42 & 0.33  & 0.30 & - &  - & -  & - & - & - & -\\
					Self-Monitoring~\cite{ma2019self} & 5.67 & 0.59 & 0.48  & 0.35  &-  &-  &- &- &373 &0.61 &0.02\\
					PTA~\cite{landi2019perceive}  & 6.17 &  0.47 & 0.40  & 0.36 & - &  - & -  & - & - & - & -\\
					Reinforced Cross-Modal~\cite{wang2019reinforced} & 6.12 &  0.50 & 0.43  & 0.38 & 4.21 &  0.67 & 0.61  & 0.59 &358 &0.63 &0.02 \\
					Regretful Agent~\cite{ma2019regretful} &  5.69 &   0.48 & 0.56  & 0.40  &-  &-  &- &- &13.69 &0.48 &0.40\\
					FAST~\cite{ke2019tactical} &  5.14 &   - & 0.54  & 0.41  &-  &-  &- &- & 196.53 & 0.61 &0.03\\
					EGP~\cite{deng2020evolving} & 5.34 &  0.61 & 0.53  & 0.42 & - &  - & -  & - & - & - & - \\
					ALTR~\cite{huang2019transferable} & 5.49 &  - & 0.48  & 0.45 & - &  - & -  & - &- &- &-\\
					Environmental Dropout~\cite{tan2019learning} & 5.23 &  0.59 & 0.51  & 0.47 & 3.97 &  0.70 & 0.64  & 0.61 &687 &0.69 &0.01\\
					SERL~\cite{wang2020soft} & 5.63 &  0.61 & 0.53  & 0.49 & - &  - & -  & - & - & - & -\\
					OAAM~\cite{qi2020object} & - &  0.61 & 0.53  & 0.50  & - &  - & -  & - & - & - & -\\
					CMG-AAL~\cite{zhang2020language} & 4.61 &  - & 0.57  & 0.50 & - &  - & -  & - & - & - & -\\
					AuxRN~\cite{zhu2020vision} & 5.15 &  0.62 & 0.55 & 0.51 & 3.69 & 0.75 & 0.68 
					& 0.65 & \textbf{41} &\textbf{0.71} &\textbf{0.21} \\
					DASA~\cite{sun2021depth} & 5.11&  - & 0.54  & 0.52 & - &  - & -  & - & - & - & -\\
					RelGraph~\cite{hong2020language} & 4.75 &  - & 0.55  & 0.52  & - &  - & -  & - & - & - & - \\
					ORIST$^{\star}$~\cite{DBLP:journals/corr/abs-2104-04167} & 5.10 &  - & 0.57  & 0.52  & - &  - & -  & - & - & - & - \\
					PRESS$^{\star}$~\cite{li2019robust} & 4.53 &  - & 0.57  & 0.53 & - &  - & -  & - & - & - & - \\
					PRRVALENT$^{\star}$~\cite{DBLP:conf/cvpr/HaoLLCG20} & 4.53 &  - & 0.57  & 0.53 & - &  - & -  & - & - & - & - \\
					NvEM~\cite{DBLP:journals/corr/abs-2107-07201} & 4.37 &  - & 0.58  & 0.54 & - &  - & -  & - & - & - & - \\
					SSM~\cite{wang2021structured} & 4.57 &  0.70 & 0.61  & 0.46 & - &  - & -  & - & - & - & - \\
					VLN$\circlearrowright$BERT$^{\star}$~\cite{hong2020recurrent}& 4.09 &  - & 0.63  & 0.57  & - &  - & -  & - & - & - & - \\
					Active Exploration~\cite{wang2020active} & 4.33 &  \textbf{0.71 }& 0.60  & 0.41 & \textbf{3.30} &  \textbf{0.77} & \textbf{0.70 } & \textbf{0.68} & 176.2 & \textbf{0.71} & 0.05\\
					REM(Based on VLN$\circlearrowright$BERT)~\cite{DBLP:journals/corr/abs-2106-07876} & \textbf{3.87} &  - &\textbf{0.65} & \textbf{0.59}  & - &  - & -  & - & - & - & - \\
					\hline
					
				\end{tabular}
			} 
		}
	\end{center}
	\caption{
		Leaderboard results comparing  all models on test split in unseen environments. We compare three training settings:  Single  Run  (without seeing unseen environments),  Pre-explore  (finetuning in unseen environments), and  Beam  Search(comparing success rate regardless of TL and SPL). The primary metric for Single Run and Pre-explore is SPL, while the primary metric for Beam Search is the success rate (SR).  We only report two decimals due to the precision limit of the leaderboard. $\star$ means using pre-training method.
	}
	\vspace{-9pt}
	\label{table:result}
\end{table*}
\subsection{Variations of R2R task}

\subsubsection{Room-for-Room, Room-6-Room, Room-8-Room}
Since all R2R reference paths in the data generation process are the shortest path to the target., there is a certain contradiction between following on instructions and reaching the destination. For properly evaluating consistency, dataset is supposed to be larger and has more diverse reference paths. 
To address the lack of path variety,~\citet{DBLP:conf/acl/JainMKVIB19} propose a data augmentation strategy that generates longer and more tortuous paths without additional human or low-fidelity machine annotation, which is more challenging dataset named Room-for-Room (R4R). Inspired by R4R, \citet{zhu2020babywalk} create two datasets of longer navigation tasks, Room-6-Room (R6R) and Room-8-Room (R8R).

\subsubsection{Room-to-Room with Continuous Environment}
In R2R task, the agent can only move on a fixed traversable nodes, rather than freely in the environment. Therefore,~\citet{DBLP:conf/eccv/KrantzWMBL20} propose \textit{Vision-and-Language Navigation in Continuous Environments} (VLN-CE) task, where the agent can move freely in the environment rather than transporting between pre-defined navigable nodes. This task setting introduces many challenges that are ignored in previous work. 

For solving this problem,~\citet{DBLP:journals/corr/abs-2012-05292} propose a modular approach using topological maps, since the conventional end-to-end approaches are struggle in freely traversable environments. They decomposed VLN-CE tasks in two stages: planning and control. During the exploration, topological map representation is built and used on navigation plan stage. The local controller receives the navigation plan and generates low-level discrete action.
\citet{DBLP:journals/corr/abs-2012-05446} point out that different robots often equipped with various camera configurations, these differences make it difficult to directly transfer the learned navigation skill between robots. Consequently, a generalization strategy is proposed for visual perception based on meta-learning, which enables the agent to quickly adapt to a new camera configuration through few-shot learning.

\subsubsection{Room-Across-Room, Cross Lingual R2R}
\citet{DBLP:conf/emnlp/KuAPIB20} introduce Room-Across-Room (RxR), a multilingual extension of R2R task containg English, Hindi, and Telugu. The size, scope and detail of RxR has dramatically expands comparing with R2R dataset, which contains 126K instructions covering 16.5K sampled guide paths and 126K human follower demonstration paths. Every instruction is accompanied by a follower demonstration, including a perspective camera pose trace that shows a play-by-play account of how a human interpreted the instructions given their position and progress through the path. 

~\citet{DBLP:journals/corr/abs-1910-11301} collect a Cross Lingual R2R dataset, which extends the original benchmark with corresponding Chinese instructions. Then they propose a principled meta-learning method that dynamically utilizes the augmented machine translation data for zero-shot cross-lingual VLN. 

\subsection{Street View Navigation}
Existing works mainly focus on simple visual input, and the environments are indoor scenes mostly. Actually, the complexity and diversity of the visual input in these environments are limited, since the challenges of language and vision have been simplified. Therefore, many researchers are interested in outdoor navigation based on Google Street View. 

\subsubsection{Touchdown}
\citet{DBLP:journals/corr/abs-1811-12354} propose Touchdown, a dataset for natural language navigation and spatial reasoning using real-life visual observations. They define two tasks (\ie, navigation and spatial description resolution) that require the agent to address a diverse set of reasoning and learning challenges. This is the first large scale outdoor VLN task. In this task, the agent receives a 360\textdegree RGB panoramic images when it moves to every navigable point, which is connected with undirected navigation graph. The environment includes $29{,}641$ panoramic images and $61{,}319$ edges from New York City. 

~\citet{DBLP:journals/corr/abs-2009-13112} focus on endowing agent with recognizing ability and stopping at the correct location in complicated outdoor environments. They introduce \textit{Learning to Stop} module to address the issues, which is a simple and model-agnostic module that can be facilely added into other models to improve their navigation performances.

For facilitating the experiment, \citet{DBLP:journals/corr/abs-2007-00229} divide the original StreetLearn dataset into a small part, \ie, Manh-50, which mainly covers the Manhattan area with 31K training data.
In addition, Multimodal Text Style Transfer learning approach is proposed to generate style-modified instructions for external resources and address the data scarcity issue.

\subsubsection{StreetLearn}\label{streetlearn}
For improving the end-to-end outdoor VLN research, 
\citet{DBLP:journals/corr/abs-1903-01292} present StreetLearn task based on Google Street View, which is a egocentric photo-realistic interactive outdoor navigation task. 
~\citet{2020Retouchdown} are publicly releasing the 29k raw Street View panoramic images needed for Touchdown. Evaluation metrics for this task are Task Completion, Shortest-path distance, Success weighted by Edit Distance, Normalized Dynamic Time Warping and Success weighted Dynamic Time Warping. Due to the length limitation of this paper, details of these metrics can be found in~\cite{2020Retouchdown}.
~\citet{DBLP:conf/nips/MirowskiGMHATSK18} use deep reinforcement learning method to address outdoor city navigation issue. Based on the images and connectivity on Google Street View, a duel pathway navigation model is proposed with interactive environment.

\subsubsection{Other outdoor navigation tasks}
\begin{itemize}
	\item \textbf{StreetNav.} StreetNav is a extension of StreetLearn proposed by~\citet{DBLP:conf/aaai/HermannMMBAH20}. The main difference is in StreetNav, driving instructions from Google Maps by randomly sampling start and goal positions are added to dataset, which likes autonomous driving environment. 
	\item \textbf{Talk to Nav.}
	\citet{DBLP:journals/ijcv/VasudevanDG21} develop an interactive visual navigation environment based on Google Street View named Talk to Nav dataset with 10,714 routes.
	They also design an effective model to create large-scale navigational instructions over long-range city environments.
	\item \textbf{Street View.}
	Based on Google Street View, ~\citet{cirik2018following} sample 100k routes in 100 regions in 10 U.S cities to build an outdoor instructions following task environment. They argue outdoor navigation is more challenging task because the outdoor environment is more chaotic and objects are more diverse.
	\item \textbf{RUN.} 
	Aiming to explaining navigation instructions based on real, dense, and urban map,~\citet{DBLP:conf/emnlp/Paz-ArgamanT19} propose \textit{Realistic Urban Navigation} (RUN) task, which has 2,515 instructions annotated by Amazon Mechanical Turk works.
	\item \textbf{ARRAMON.} 
	\citet{DBLP:conf/emnlp/KimZBTB20} propose ARRAMON task, which contains two sub-tasks: object collecting and object referring expression comprehension. During this task, the agent is required to find and collect different target objects one by one through natural language instruction-based navigation in a complex synthetic outdoor environment.
\end{itemize}

\section{Multi-turn Tasks}\label{sec:mul}
Prior works require the agent to navigate in an environment with a single-turn instruction. Recently, there is surge of dialog-enabled interactive assistant, where the interaction is often a multi turn process. For multi-turn tasks, instructions will be given by a guide to the agent in several turns, until the agent has reached the specified goal. According to whether the agent can question the instruction or not, we subdivided multi-turn tasks into: \textit{Passive} and \textit{Interactive} tasks. 

\subsection{Passive task}
In passive task, instructions are given to the agent in stages, where the information contained in the instructions is often sufficient and unambiguous, so the agent only needs to understand the meanings and move to the target position. Order instructions for this type of task is different for this problem. 

\subsubsection{CEREALBAR}
CEREALBAR~\citep{DBLP:conf/emnlp/SuhrYSYKMZA19} simulates a scenario, where a leader and a follower collaboratively select cards to earn points. If valid set of card is collected, the players can earn a point. The leader has full observability on the map, and are plausible to plan the path for the follower and gives instructions to direct the way. The follower can only see a first-person view of the environment, and cannot respond to the leader. The performance is measured by correct execution of instructions and the overall game reward.

To solve this task,~\citet{DBLP:conf/emnlp/SuhrYSYKMZA19} introduce a learning approach focusing on recovery from cascading errors between the sequential instructions, and modeling methods to explicitly reason about instructions with multiple goals.
\subsubsection{HANNA}
HANNA~\citep{DBLP:conf/emnlp/NguyenD19} defines a task for a human to find objects in an indoor environment. In this process, the human can ask a mobile agent via natural language. The task is a high-level command (“find [object(s)]"), modeling the general case when the requester does not need know how to accomplish a task when requesting it. HANNA uses the Matterport3D simulator to photo-realistically emulate a first-person view while navigating in indoor environments.

To address the HANNA problem, \citet{DBLP:conf/emnlp/NguyenD19} develop a hierarchical decision model via a memory-augmented neural agent. Imitation learning is used to teach the agent to avoid past mistakes and make future progress.

\subsubsection{VNLA}
Based on Matterport3D simulator, \citet{DBLP:conf/cvpr/NguyenDBD19} introduces the \textit{Vision-based Navigation with Language-based Assistance} (VNLA) task, where an agent has visual perception in a photo-realistic indoor environment and is guided to find objects in the environment via instructions.

Furthermore, they construct ASKNAV, a dataset for the VNLA task. After filtering out labels that occur less than five times, 289 object labels and 26 room labels are obtained. The data is defined as a tuple (environment, start pose, goal viewpoints, end-goal).~\citet{DBLP:conf/cvpr/NguyenDBD19} develop a general framework with imitation learning to extend the framework for indirect intervention.

\subsection{Interactive task}
In interactive task, the human guide and the agent are usually cooperative, and the agent can ask the person if it encounters a situation of information insufficiency or ambiguity. In this interactive manner, the agent can continuously acquire information to complete the navigation task.

\subsubsection{Talk The Walk}
\citet{DBLP:journals/corr/abs-1807-03367} introduces a navigation task for tourist service and constructs a large-scale dialogue dataset named ``Talk The Walk”. The tourist is located in a virtual 2D grid environment in New York City and has 360-views of the neighborhood blocks. The guide has an abstracted semantic map of the blocks and the target location is unambiguously shown to the guide from the start. A guide aims to interact with a “tourist”via natural language to help the latter navigate towards the correct location. 

\citet{DBLP:journals/corr/abs-1807-03367} decouples the task of tourist localization from the task and introduces a Masked Attention for Spatial Convolutions (MASC) mechanism to ground the instructions from tourist into the guide's map.

\subsubsection{Navigation from Dialog History}
Robot navigation in real environments are expected to use natural language to communication and understand human's meaning. To study this challenge, \citet{DBLP:conf/corl/ThomasonMCZ19} introduce CVDN, a human to human dialogues situated in simulated, photo-realistic home environments. For training agent to search a goal location in an environment, they define the \textit{Navigation from Dialog History} (NDH) task. In NDH, an agent is learned to predict navigation actions and find an object given a dialog history between human in previously unseen environments. NDH task can be used for training agents in navigation, question asking, and question answering problems. \citet{DBLP:conf/corl/ThomasonMCZ19} formulated a Seq2Seq model to encode an entire dialog history to address the NDH task, whose outputs are actions in the environment.
\citet{DBLP:conf/cvpr/0004ZZLJCL20} propose a \textit{Cross-modal Memory Network} (CMN) to remember the historical information. CMN uses a language memory module to learn latent relationships between the current conversation and a dialog history and uses a visual memory module to learns to associate the current visual views and the previous navigation actions. \citet{roman-roman-etal-2020-rmm} divide multi-turn NDH task into three sub-tasks: question generation, question answering, and navigation and implement them with three Seq2Seq models, i.e., a Questioner, a Guide and a Navigator. The progress agents are trained with reward signals from navigation actions, question and answer generation.

\subsubsection{Just Ask}
Base on the R2R task~\citep{DBLP:conf/cvpr/AndersonWTB0S0G18}, \citet{chi2020just} develop an interactive learning framework to allow the agent to ask for users' help when needed. And they use reinforcement learning with a proposed reward shaping term as the initial model, which enables the agent to ask questions only when necessary.

\subsubsection{RobotSlang}
\citet{DBLP:journals/corr/abs-2010-12639} decouple the cooperative communication task into \textit{Localization from Dialog History} (LDH) and NDH task, where a driver agent must localize in the global map or navigate towards the next target object according to the visual observations and the dialog with the commander. Different from other tasks, the environment and the agent are physical, while the sensor observations are camera RGB images. Moreover, \citet{DBLP:journals/corr/abs-2010-12639} evaluate human performance on the LDH task and create an initial, Seq2Seq model for the NDH task.

\section{Conclusions and Discussions}\label{sec:con}
Although the application of deep learning has made gratifying progress on the VLN approaches, it is worth noting that there are still limitations in many aspects. In this section, we outline some of limitations and opportunities in future research directions. Specifically, we will discuss limitations of poor generalization, lack of high-quality data and weak interactivity of the environment. Further, we discuss the future direction of incorporating knowledge and simulation to reality.

\subsection{Limitations}

\subsubsection{Poor Generalization}  The problem is mainly reflected in two aspects: one is that when trained VLN model is transferred from seen scenes to unseen scenes, the model performance will drop significantly; besides, when the model is transferred from one task to another VLN task, the performance will also decrease. The possible reason is that the data distributions in different environments vary greatly; hence, the navigation model trained in one environment is difficult to adapt to other environments. Another explanation may be that there is a certain degree of over-fitting in the supervised learning method. 
	
\subsubsection{Lack of high-quality data} A lot of work~\cite{fried2018speaker,DBLP:journals/corr/abs-2106-07876} has demonstrated that through data augmentation, generating more training data can improve the performance of VLN model. For example, ~\citet{DBLP:conf/cvpr/WijmansDMDGLEPB19} have proved that the point cloud perception can help an EQA agent find its target. However, compared with the real high-quality data, the artificially generated data is still very limited to the model. Unfortunately, collecting real data is still a costly and labor-intensive work, since the methods to collect photo-realistic 3D-scene datasets require scanning real buildings by means of expensive photogrammetry devices~\cite{mikhail2001introduction} such as Matterport 3D scanner, Meshroom or even mobile 3D scanning applications. 

\subsubsection{Weak interactivity with the environment} In the current VLN datasets, the interaction between the agent and the environment is simplified, only involving basic operations such as opening a microwave oven, moving an object, or reaching a designated location. The movement action that can be taken are restricted to a specific range (such as moving forward, turning left and turning right, or teleporting between navigable point). Although more complex interaction forms such as ``bring me the pillow next to the stool" have appeared in some latest research, it is obviously far from the interaction and movement forms of the real environment, because they simplify many physical constraints in the real environment. In the future, a simulator with advanced physics features such as cloth, fluid and soft-body physics will improve the realism of intelligent VLN agent.

\subsection{Opportunities}

\subsubsection{Incorporating knowledge}
Knowledge makes human smarter, for instance, we can summarize experience from daily life as knowledge and use it in the future. Intuitively, knowledge will imporve the model performance in most of AI tasks, which has been proven effective recently~\cite{DBLP:journals/corr/abs-2002-11310,yang2018visual}. Therefore, how to introduce knowledge into the VLN model and make it work effectively is a problem that researchers need to think about in the future. Some possible ways are as following:
\begin{itemize}
	\item \textbf{Algorithmic knowledge.} Many algorithms in machine learning can accumulate knowledge through continuous training such as the knowledge about environmental dynamics, spatial representation or physical effects of actions, so that they can better adapt to new environments. \citet{DBLP:journals/corr/abs-2012-05446} used the Model-Agnostic Meta-Learning (MAML) algorithms to improve model generalization in task VLN-CE, which enables the agent to get adaptive quickly to a new camera configuration with a few shots. Another work~\cite{DBLP:conf/cvpr/WortsmanERFM19} adopted MAML to learn a self-supervised loss, which can be used in new environments directly without extra training.
	Besides, lifelong learning~\cite{DBLP:journals/ral/LiuXS21}, transfer learning~\cite{huang2019transferable}, curriculum learning~\cite{zhu2020babywalk} and external prior knowledge~\cite{DBLP:conf/iclr/YangWFGM19, DBLP:conf/kse/NguyenNL19} are introduced to help solving VLN tasks.
	
	\item \textbf{External knowledge base.} Humans often need the help of commonsense when solving similar problems. For example, with the commonsense knowledge that a toilet is typically found in the washroom, the agent shouldn’t waste time in the living room to find a toilet.  Moreover, such knowledge is likely to still hold in previously unseen environments, making it possible to achieving better generalization ability. Some existing pre-training models can play the role of some common sense knowledge bases, such as some image-text pairs, but they are still far from the ability to have common sense reasoning. How to make the agent have common sense knowledge and make it reason on the basis is a very critical problem.
	
	\item \textbf{Cognitive architecture.} Humans can perform tasks like VLN excellently, so some principles in cognitive science and human brain can be used for heuristic model building. For example, memory can enhance the reasoning ability of VLN model. In brain areas, such as the hippocampus, have intertwined roles in spatial memory and other cognitive functions. Inspired by this,  \citet{hu2021vision} use fusion adaptive resonance theory (ART) networks to model spatial representation and related memory to enhance the visual navigation model. Other useful cognitive theories like Soar, ACT-R are powerful as well, and no work based on them until now.

\end{itemize}

\subsubsection{From simulation to reality}
It is promising for a physical agent to complete VLN tasks in the real world, since there are many practical application scenarios. For instance, in search-and-rescue tasks, robots might using vision clues to search and language to interact with with human victims~\cite{tadokoro2009rescue}. 
At present, the research on VLN task is still in the simulated environment. The advantage of the simulation platform is that it can easily validate the model, and it is convenient and quick to use. However, some important physical constraints in many real-world cases cannot be characterized well by simulators and have been discarded from consideration. Therefore, there is still a large gap between the simulated environments and the real world, hindering the transfer of progress made on simulation platforms to real world scenarios. Some progress has been made in this direction. For example,~\citet{anderson2020sim} have used a TurtleBot2 mobile robot equipped with a 2D laser scanner and a 360\textdegree \ RGB camera to finish a VLN task. In the future, we believe that building a simulated environment integrated with more realistic physics features, such as tactile perception~\cite{bhirangi2021reskin}, human-level motor control~\cite{smith2020constraining} and audio inputs~\cite{chen2019audio} will improve the performance of VLN model in real world.

Besides, training VLN models usually need intensive computation resources, which makes it prohibitive to deploy such models on portable devices, such as small drones~\cite{DBLP:conf/rss/BlukisBBKA18}. One direction to solve the problem is to design more lightweight network structure,  where operations or layers with low computing load and low memory consumption are included. For example, the model can be compressed by quantizing the network weight or being distilled to a light model to reduce the complexity. Other possible solutions could be model pruning method and Huffman coding for weights of the model.

\subsection{Conclusions}
Recent advances in computer vision, natural language processing and photo-realistic simulators have been a key driver of progress in VLN research.
In this survey, we have comprehensively reviewed existing VLN tasks, introduced typical datasets and simulators, and thoroughly summarized newest research progress. We argue that the most important contribution is the new taxonomy and believe that our taxonomy will help researchers to categorize future tasks, better understand the remaining unresolved problems, and be able to choose the most suitable VLN simulators for their research tasks. Furthermore, we analysis the current limiations exist in VLN field and point out a promising direction in the future.
 
\section*{Acknowledgment}
The work described in this paper was sponsored in part by the National Natural Science Foundation of China under Grant No. 62103420 and 62103428 , the Natural Science Fund of Hunan Province under Grant No. 2021JJ40702.

\ifCLASSOPTIONcaptionsoff
  \newpage
\fi



%
%
%
\footnotesize
\bibliographystyle{IEEEtranN}
\bibliography{reference}
\end{document}